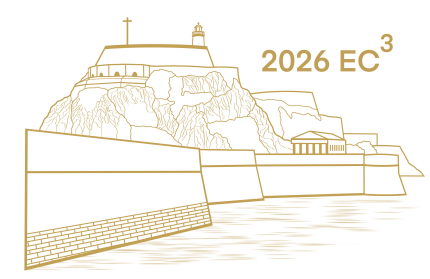

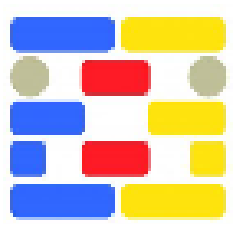

# Benchmarking Deep Learning Approaches for AEC Engineering Drawing Layout Detection and Information Extraction

Tianyang Huang[1,2], Alessio Lombardi[1], Ahmed Elnagar[1], Ahmed Zalouk[2], George Paul[1], Sepehr Najjarpour[1]
Arvid Sigurdsson[1], Khalid Ismail[2], Mohamed Ragab[2], Edlira Vakaj[2]
[1]Buro Happold Ltd., London, United Kingdom
[2]Birmingham City University, Birmingham, United Kingdom

## Abstract

Information Extraction (IE) from Architecture, Engineering, and Construction (AEC) drawings remains hindered by manual inefficiency, while Layout Detection—a vital "middleware" organizing graphical and textual hierarchies—is underexplored. General document layout models, optimized for text-centric content, lack validation on engineering drawings. This study constructs a custom AEC-specific layouts dataset and benchmarks five deep learning architectures. RF-DETR achieves state-of-the-art performance with an $mAP_{50}$ of 0.949, while the Vision-Language Model Qwen3-VL attains a leading F1-score of 0.911. Conversely, models pre-trained on general document datasets suffer from "domain interference", causing performance degradation. This establishes a robust technical foundation for automated IE in AEC.

## Introduction

In the Architecture, Engineering, and Construction (AEC) industry, drawings are still a common medium for information exchange. Efficient Information Extraction (IE) is essential for downstream tasks such as building refurbishment, Quality Assurance (QA), and digital twin development. In contemporary practice, however, IE from drawings remains heavily reliant on manual effort, which is labor-intensive and error-prone. Automating drawing interpretation is thus an urgent industrial priority.

Information within an engineering drawing can be systematically categorized into three hierarchical layers: the *Metadata Layer* (titleblocks), the *Layout Layer* (spatial arrangement), and the *Main Drawing Content Layer* (graphical entities) (shown in Figure 1). Serving as the vital "middleware" between these two, the *Layout Layer* defines the logical relationships and hierarchical structure between disparate elements, such as drawing areas, notes, and titleblocks.

Previous efforts have primarily focused on isolated layers of this hierarchy, such as automated metadata extraction from titleblocks (Lombardi et al., 2025) or localized symbol detection (Khan et al., 2024; Jamieson et al., 2025). While advancing specific tasks, these studies often overlook the overall engineering drawing layout. This research gap is critical because the layout provides the semantic framework necessary for holistic understanding; for instance, the meaning of a symbol often depends on its functional orientation within the layout context. Without this structural link, isolated IE loses the global logic required for complex engineering reasoning.

While Document AI has produced powerful Document Layout Analysis (DLA) models like LayoutLM (Xu et al., 2020, 2021; Huang et al., 2022) and DocLayout-YOLO (Zhao et al., 2024b), these are primarily optimized for general text-centric documents such as invoices or academic papers. Due to the unique characteristics of engineering drawings—such as extreme aspect ratios, high-resolution requirements, and sparse graphical density—the performance of general-purpose Document AI models in this domain has not been systematically benchmarked or validated.

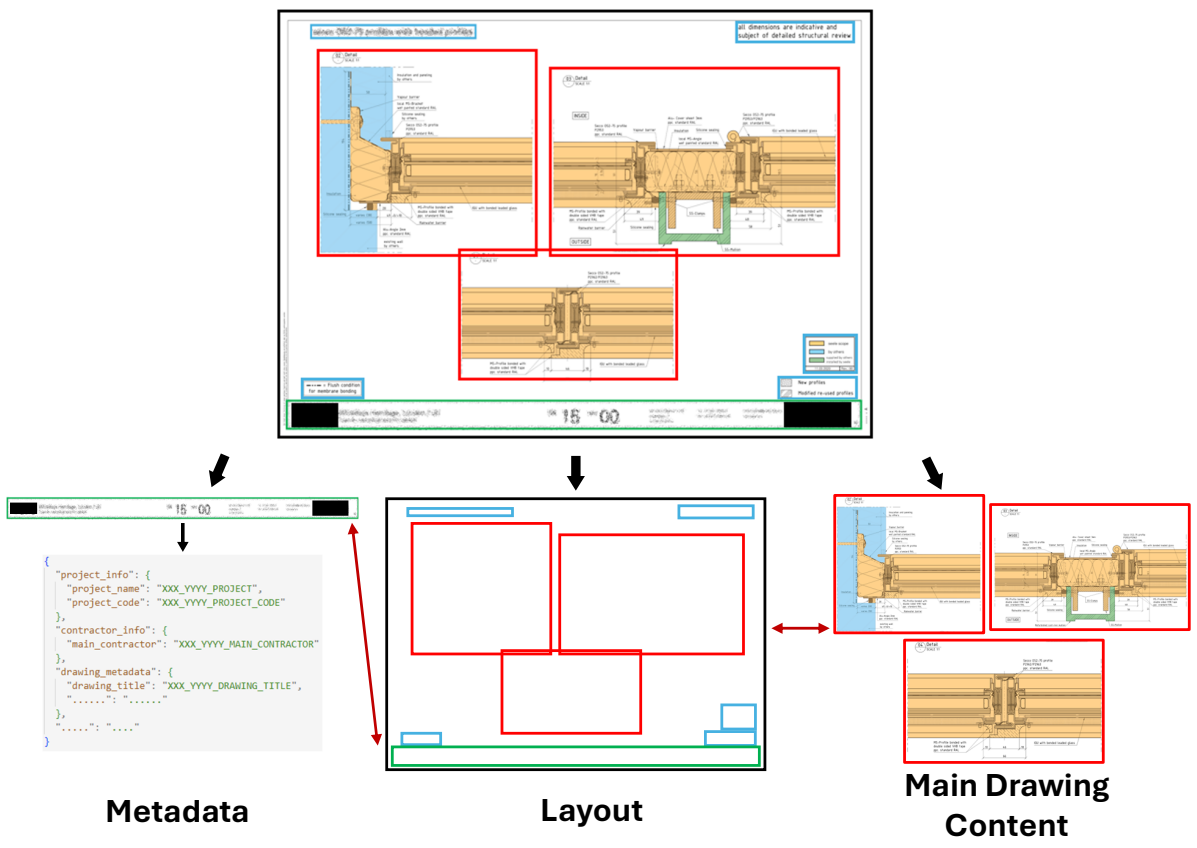

*Figure 1: Hierarchical layers of information in AEC drawings*

To fill this research gap, we collaborated with industry experts to construct the *Facade Engineering Layout Dataset*, comprising 551 high-quality annotated drawings. Based on this dataset, we conducted a comprehensive benchmark of state-of-the-art models, including Faster R-CNN (Ren et al., 2017), YOLOv12 (Tian et al., 2025), RF-DETR (Robinson et al., 2026), Qwen3-VL (Bai et al., 2025) and DocLayout-YOLO, evaluating their efficacy in engineering drawing layout detection. Additionally, an ablation study on DocLayout-YOLO investigates whether general DLA architectural designs and pre-trained weights can be effectively transferred to the AEC domain.

The primary contributions of this research are as follows.

1. We identify the importance of Layout Detection in the AEC domain as a prerequisite for holistic drawing understanding.
2. We provide an extensive benchmark of multiple deep learning architectures, analyzing their strengths and limitations when dealing with large-scale, high-complexity engineering documents.
3. We reveal a significant "domain interference" effect through the ablation study, demonstrating that specialized architectural designs and pre-trained weights from general document domains can act as negative priors for AEC drawings.

Our findings establish a robust technical pathway for performing detailed IE and QA on engineering drawings in AEC projects.

## Related Work

### Information Extraction in Engineering Drawings

Engineering drawings encapsulate high-density geometric and semantic data, necessitating their transformation into structured formats for industrial digitalization. Current research predominantly bifurcates into two domains: Metadata Extraction from titleblocks and Main Drawing Content Analysis.

*Metadata Extraction from Titleblocks*

Early heuristic and OCR-based methods (Kashevnik et al., 2023) often lack robustness against non-standard layouts or scanning artifacts. Recent advancements have shifted toward Transformer-based end-to-end detection. Microsoft's Donut (Kim et al., 2022) bypassed traditional OCR errors by employing a joint visual encoder and language decoder. Furthermore, Lombardi et al. (2025) integrated Faster R-CNN with GPT-4o to parse titleblocks in AEC drawings into structured key-value pairs. They achieved 86% key retrieval and 95% value-matching accuracy, demonstrating Multimodal Large Language Model's potential for semi-structured AEC documents.

*Main Drawing Content Analysis*

In contrast to titleblocks, information within the main drawing area is dispersed and topologically complex. Zhang et al. (2023) addressed this by modeling mechanical drawings as graph structures, utilizing GraphSAGE to classify geometric contours and annotations. Khan et al. (2026) proposed a two-stage hybrid framework using YOLOv11-OBB to identify text regions (e.g., GD&T, materials, surface roughness), followed by a finetuned Donut model for structured JSON extraction. To extract main drawing content from high-resolution AEC drawings, current studies frequently employ tiling strategies—partitioning massive images into smaller patches to ensure compatibility with Vision Transformers or CNN-based detectors (Jamieson, 2024; Carrara et al., 2025).

*Ontology-based floor plan analysis (DAnO)*

The work of Schönfelder and König (2025) proposed the *Drawing Analysis Ontology* (DAnO) as an ontology-driven framework for automatic floor plan analysis, focusing primarily on architectural plans. It provides a domain model to represent drawing elements and their semantic relationships, and it is used to support reasoning tasks such as aggregation and consistency checking on floorplan content. In typical DAnO use cases, each drawing sheet contains a single main plan view, and the ontology targets the interpretation of this view rather than the explicit modelling of multiple layouts within a sheet.

*Layout Detection: The Missing Link*

Existing AEC "layout" research is primarily restricted to architectural floor plans, focusing on semantic segmentation of object elements (e.g. walls, doors) within individual views (Heras et al., 2014). These approaches fail to address the holistic sheet-level layout of generic AEC drawings, where heterogeneous views—such as plans, sections, and elevations—coexist. To bridge this gap, we propose multi-view layout recognition as a strategic "middleware" stage that facilitates holistic data extraction through the following benefits:

**Semantic Contextualization**: Layout detection identifies the spatial distribution of drawing areas, which is critical as identical symbols carry distinct semantic meanings depending on their viewport (e.g., plan vs. section). Without this context, isolated symbol detection loses essential semantic associations.

**Optimizing High-Resolution Processing**: Unlike "blind" tiling techniques that cause semantic fragmentation by severing continuous engineering entities, prioritizing layout detection identifies information-dense "Regions of Interest" (RoI). This facilitates semantic-aware tiling, ensuring high-resolution crops maintain logical boundaries and necessary global context for downstream extraction tasks.

### General Document Layout Analysis

While layout analysis remains under-explored in the AEC sector, general Document Layout Analysis (DLA) has transitioned from early heuristic rules (Namboodiri and Jain, 2007) to sophisticated deep learning paradigms. This shift was catalyzed by large-scale benchmarks such as PubLayNet (Zhong et al., 2019) and the structurally diverse DocLayNet (Pfitzmann et al., 2022); the latter has become the primary benchmark in the field due to its accurate and detailed manual labels. Currently, the field is dominated by multimodal Transformers, notably the LayoutLM series (Xu et al., 2020, 2021; Huang et al., 2022), which integrates visual and positional features for SOTA performance. To meet industrial throughput requirements, models like DocLayout-YOLO (Zhao et al., 2024b) have emerged, utilizing a G2L_CRM architecture to achieve competitive accuracy with significantly faster inference speeds. Recent research has further matured toward advanced structural understanding, including reading path reconstruction (Wang et al., 2021) and graph-based document structure analysis (Chen et al., 2025).

*The Research Gap: Unique Challenges of AEC Drawings*

Despite the maturation of general DLA, research specifically tailored to AEC drawings remains sparse. Models optimized for general documents cannot be directly migrated due to data scarcity and significant domain disparity. Unlike text-centric documents, AEC drawings feature massive formats, a dominance of non-textual symbols, and complex topological interdependencies that require specialized handling.

To address these challenges, this study constructs a high-quality *Facade Engineering Layout Dataset* and provides a comprehensive benchmark of SOTA models on this specialized domain. Moreover, an ablation study of DocLayout-YOLO is conducted to systematically evaluate how its specialized G2L_CRM architecture and large-scale document pre-training weights influence performance when transferred to the unique graphical landscape of AEC engineering drawings.

## Methodology

In this section, we outline our methodology for benchmarking engineering drawing layout detection. We first introduce the curation of the *Facade Engineering Layout Dataset* (FELD). This is followed by our selection of candidate models and the standardized experimental framework used for evaluation.

### Dataset and Ontology Overview

To support the benchmarking experiments, this study curated the FELD dataset. This dataset consists of 551 facade drawing images derived from 10 real-world construction projects, comprising nine common drawing layout elements. A comprehensive overview of the dataset is provided in Table 1 below.

*Table 1: FELD Dataset Overview*

| Category Name | Image Counts | Annotation Counts |
|---|---|---|
| Titleblock | 551 | 604 |
| Titleblock; Revision Table | 425 | 425 |
| Legend | 54 | 91 |
| Notes | 145 | 288 |
| Layout; Elevation | 56 | 60 |
| Layout; Section; Horizontal | 239 | 321 |
| Layout; Section; Vertical | 307 | 376 |
| Layout; Section; GA | 23 | 37 |
| Layout; Plan; GA | 21 | 26 |
| **All Categories** | **551** | **2228** |

**GA**: General arrangement

Prior to annotation, we first collaborated with facade domain experts to define an AEC Drawing Information Representation Ontology (ADIRO); the ontology is illustrated in Figure 2 below. This ontology is intended to delineate the spatial structure and semantic connotations of the drawings through a hierarchical logical framework. ADIRO is tailored to the needs of IE from engineering drawings, rather than directly adopting existing building ontologies or classification schemes. Existing models and ontologies (e.g., IFC, ifcOWL, Uniclass, Brick, BOT) are designed primarily for BIM authoring, lifecycle asset management, or building automation, and therefore optimize for comprehensive domain coverage and interoperability. As a result, they tend to be large, multi-purpose and conceptually heavy, with many distinctions and constraints that are irrelevant at the level of 2D drawing analysis and information extraction. For drawing IE, we instead require a schema that is tightly aligned with visual patterns in drawings, annotation workflows, and graph-based reasoning, in addition to local spatial relations. A custom ontology allows one to minimise complexity, expose only the concepts and relations that matter for IE, downstream queries, and QA checks, and iterate rapidly as new tasks emerge. Interoperability is preservable by attaching external identifiers (such as IFC types or Uniclass codes) as attributes when required, while the core representation remains simple and learnable. We included nine of the most representative and common layout categories in ADIRO (see Table 1) and established annotation guidelines.

ADIRO is closely related to DAnO (Schönfelder and König, 2025), which defines concepts such as DrawingElement, DisplayElement, and DescriptionElement, and their relationships, enabling semantic graph-based reasoning over text–symbol–geometry relations in architectural plans. In contrast, ADIRO is designed for engineering drawings where multiple layouts commonly coexist on a single sheet. To capture this, ADIRO introduces Layout as a parent concept to DrawingElement: a layout can contain drawing elements as well as other sheet content such as tables, schedules, and textual regions, thereby explicitly modelling the hierarchical structure of multi-layout sheets. While ADIRO and DAnO partially overlap at the level of basic drawing concepts, their scope and structuring differ; we therefore consider ADIRO complementary to DAnO and plan to explore explicit linking or reuse of DAnO modules in future extensions of ADIRO, once we move beyond the layout-focused core required for the present work.

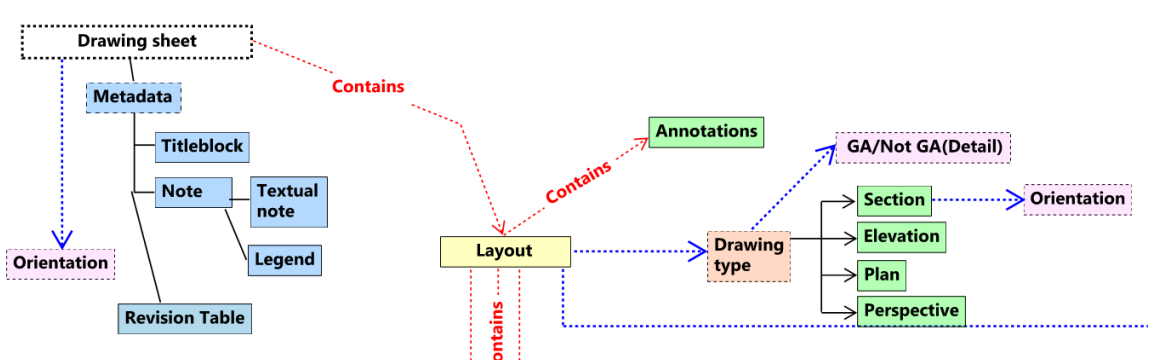

*Figure 2: Partial view of the AEC Drawing Information Representation Ontology (ADIRO)*

During the annotation phase, domain experts performed labeling which was cross-checked. Standard rasterization was performed to ensure consistency across vector and raster drawings, with all drawings at 300 DPI. Due to the varying dimensions of the originals, the dataset image dimensions range from $3509 \times 2480$ to $14042 \times 9934$ pixels.

### Evaluated Model Architectures

To comprehensively evaluate layout detection in AEC drawings, we selected five architectures representing distinct technical paradigms:

*Faster-RCNN & YOLOv12*

Faster R-CNN is employed as a robust two-stage baseline to evaluate traditional anchor-based frameworks on large-scale layouts. Conversely, YOLOv12 (m) represents the state-of-the-art in one-stage detectors, selected to assess the balance between real-time inference efficiency and detection accuracy via its Area Attention mechanism.

*RF-DETR*

RF-DETR is selected as the latest iteration of Transformer-based detection models and currently stands as a SOTA solution in the field. By reframing detection as a set prediction problem, it provides a purely end-to-end pipeline without reliance on Non-Maximum Suppression (NMS). We utilize RF-DETR to evaluate the efficacy of DETR's global attention mechanisms in modeling the intricate topological relationships between disparate drawing zones.

*Qwen3-VL*

To explore the potential of state-of-the-art multimodal models in specialized domains, we introduce Qwen3-VL. This high-performance vision-language model natively supports long-text streams up to 256K tokens and utilizes an enhanced Interleaved-MROPE (Multimodal Rotary Positional Embedding) to improve spatial modeling for large-scale documents.

By benchmarking Qwen3-VL-8B-Instruct, we evaluate whether the semantic knowledge from large-scale multimodal pre-training can be effectively transferred to the specialized domain of architectural symbolic layouts.

*DocLayout-YOLO*

Specifically engineered for general DLA, DocLayout-YOLO is included to evaluate its domain-specific architectural optimizations and whether pre-training on general document layout datasets can be effectively transferred to AEC drawings. Based on the YOLOv10m backbone, its core innovation—Global-to-Local Contextual Modeling (G2L_CRM) — enables the model to simultaneously perceive the overall logical structure of a page and fine-grained semantic details. By comparing this model with general-purpose detectors and conducting additional ablation studies, we aim to quantify the extent to which its architectural enhancements and large-scale document pre-training contribute to performance gains when applied to AEC-specific documents.

## Experimental Setup

*Hardware and Environmental Configuration*

All experiments were implemented using the PyTorch framework on the Ubuntu operating system. To ensure computational efficiency, Qwen3-VL-8B-Instruct was trained on an NC24ads-A100-v4 instance with MS-Swift framework (Zhao et al., 2024a), while Faster R-CNN, YOLOv12, RF-DETR, and DocLayout-YOLO were trained on an NC4as-T4-v3 instance.

*Model Specifications and Initialization*

To ensure a fair and comprehensive comparison, we selected specific variants of each architecture and initialized them with specialized pre-trained weights. Faster R-CNN was implemented using the ResNet-50 backbone with Feature Response Normalization (FRN). For the YOLOv12 and RF-DETR architectures, we utilized their respective medium (m) variants to balance computational cost and detection capacity. These three models were initialized with weights pre-trained on the COCO dataset. In contrast, Qwen3-VL was deployed using the 8B-Instruct version, which had undergone extensive large-scale pre-training. For DocLayout-YOLO, we utilized a version specifically pre-trained on large-scale document layout datasets, namely DocLayNet and DocSynth, to leverage its domain-specific knowledge in general DLA.

*Data Partitioning and Preprocessing*

Given the imbalanced distribution of labels in engineering drawings, we employed Multi-label Stratification (Sechidis et al., 2011) to partition the dataset into training, validation, and test sets with a ratio of 70:15:15. In the preprocessing stage, no data augmentation was applied. Images were resized while preserving their original aspect ratios to match the default input specifications of each model: $1344 \times 800$ pixels for RF-DETR, Faster R-CNN and Qwen3-VL, $640 \times 640$ pixels for YOLOv12 and DocLayout-YOLO.

For Qwen3-VL, the dataset was converted from the standard COCO JSON format into a specialized Grounding prompt format (JSONL). The model is trained to generate JSON-formatted outputs containing a list of objects with their respective category labels and 2D bounding boxes. To maximize detection performance, we implemented a dual-query strategy:

General Queries: The model is prompted to identify and locate all layout elements simultaneously (e.g., "Identify and locate all drawing layout elements...").

Per-category Queries: The model focuses on localizing a specific target class (e.g., "Locate the Legend region...").

*Training Strategy*

Training was conducted using the AdamW optimizer coupled with a Cosine Learning Rate Scheduler. A warm-up period was implemented during the first 500 iterations to stabilize the initial gradients. All architectures were initialized with pre-trained weights mentioned above and fine-tuned according to their respective default loss configurations. During model training, we implemented selective parameter freezing strategies tailored to the specific architecture of each model. Specifically, for Qwen3-VL, we employed Parameter-Efficient Fine-Tuning via LoRA (Low-Rank Adaptation) to achieve efficient adaptation.

## Evaluation Metrics

*Accuracy Metrics*

To assess the detection performance, we utilize standard COCO-style metrics and classification statistics. The core benchmarks are the mean Average Precision ($mAP$), specifically $mAP_{50}$ and $mAP_{50:95}$. The precision ($P$), recall ($R$), and F1-score are calculated at a confidence threshold of 0.25 to evaluate the balance between prediction accuracy and coverage.

*Table 2: Overall Performance Comparison of Evaluated Models on the FELD Dataset.*

| Model | Accuracy Metrics | | | | | Efficiency Metrics | | | |
|---|---|---|---|---|---|---|---|---|---|
| | $\mathbf{mAP}_{50:95}$ | $\mathbf{mAP}_{50}$ | **P** | **R** | **F1** | **Params** | **GFLOPs** | **Time** | **FPS** |
| Faster R-CNN | <u>0.779</u> | <u>0.930</u> | 0.829 | **0.924** | 0.874 | 41.39M | 187.0 | 88.0ms | 11.4 |
| RF-DETR (m) | **0.890** | **0.949** | **0.985** | 0.800 | <u>0.889</u> | 33.70M | 100.8 | <u>44.8ms</u> | <u>22.3</u> |
| YOLOv12 (m) | 0.755 | 0.871 | 0.871 | 0.793 | 0.830 | 20.06M | 67.1 | **23.1ms** | **43.3** |
| Qwen3-VL-8B | 0.668 | 0.758 | <u>0.916</u> | <u>0.906</u> | **0.911** | 8.81B | – | 6120.3ms | 0.16 |
| DocLayout-YOLO† | 0.589 | 0.675 | 0.736 | 0.602 | 0.662 | 19.97M | 68.0 | 30.0ms | 33.4 |

† Pre-trained on DocLayNet and DocSynth datasets. **P**: Precision; **R**: Recall; **F1**: F1-score; **Time**: Inference time per unit.

*Table 3: Detailed performance comparison ($mAP_{50:95}$) across specific layout categories.*

| Model | Title | Rev. | Notes | Legend | H-Sec | V-Sec | GA | Elev. | Plan |
|---|---|---|---|---|---|---|---|---|---|
| Faster R-CNN | 0.866 | <u>0.924</u> | 0.666 | 0.896 | 0.734 | 0.747 | <u>0.820</u> | 0.722 | <u>0.775</u> |
| RF-DETR (m) | **0.989** | **0.932** | **0.734** | 0.928 | **0.917** | <u>0.845</u> | **0.843** | **0.849** | **0.969** |
| YOLOv12 (m) | <u>0.952</u> | 0.886 | <u>0.684</u> | <u>0.910</u> | <u>0.757</u> | 0.804 | 0.500 | <u>0.794</u> | 0.510 |
| Qwen3-VL-8B | 0.939 | 0.903 | 0.560 | **1.000** | 0.717 | **0.856** | 0.554 | 0.501 | 0.330 |
| DocLayout-YOLO† | 0.894 | 0.713 | 0.395 | 0.720 | 0.554 | 0.684 | 0.039 | 0.505 | 0.409 |

**Title**: Titleblock; **Rev.**: Revision table; **H-Sec**: Horizontal section; **V-Sec**: Vertical section; **GA**: GA section; **Elev.**: Elevation.

† Pre-trained on DocLayNet and DocSynth.

*Model Complexity and Efficiency*

Beyond accuracy, we quantify the computational requirements and real-time performance of each model. Model complexity is measured by the number of Parameters and Floating Point Operations (FLOPs). To evaluate practical deployment feasibility, we record models' Inference Time per unit (pre-processing and post-processing time excluded) and Frames Per Second (FPS) on T4 GPU.

## Results and Analysis

This section presents a comprehensive analysis of the experimental results achieved by the candidate models on the FELD dataset. We first evaluate the overall performance of each architecture from the perspectives of both accuracy and computational efficiency, followed by an in-depth exploration of performance variations across different layout categories. Furthermore, an ablation study is conducted on DocLayout-YOLO to investigate how its specialized architectural innovations for general DLA, combined with pretrained weights from general document datasets, influence the model's performance on AEC-specific drawings.

### Overall Performance Analysis

Table 2 presents the overall performance of the candidate models on the FELD dataset. Throughout this paper, the best-performing results for each metric are highlighted in bold, while the second-best are indicated with an underline.

First, RF-DETR (m) achieves the SOTA performance in detection accuracy, reaching a peak $mAP_{50}$ of 0.949, $mAP_{50:95}$ of 0.890, and a Precision of 0.985. This result demonstrates the superiority of the global attention mechanism within the Transformer architecture when processing large-scale drawings. In contrast, the traditional Faster R-CNN exhibits significant robustness, ranking second in $mAP_{50}$ (0.930) and $mAP_{50:95}$ (0.779) while achieving the highest Recall (0.924). This validates the continued efficacy of the Region Proposal Network in identifying regular geometric layouts prevalent in AEC drawings. However, its inference speed is limited to 11.4 FPS. Regarding YOLOv12 (m), this model remains the most efficient architecture, achieving the highest frame rate of 43.3 FPS and a leading inference time of 23.1ms while its Precision (0.871) remains competitive.

The vision-language model Qwen3-VL-8B demonstrates a unique performance profile influenced by its specific evaluation methodology. During inference, the model performs per-category queries. Following the standard convention for VLM-based grounding tasks, the detection confidence for all outputs is fixed at 1.0. This could negatively impact metrics such as $mAP$, which rely on sorting detections to calculate precision-recall curves.

Despite this inherent disadvantage in ranking-based metrics, the model achieves the highest F1-score (0.911) and maintains strong performance with the second-highest Precision (0.916) and Recall (0.906). This suggests that the extensive multimodal knowledge and vision-language alignment embedded in Qwen3-VL provide superior semantic classification consistency. However, its localization precision ($mAP_{50:95}$ of 0.668) still lags behind specialized models. Moreover, its significant 8B parameter count leads to extreme latency, with an inference time of 6120.3ms per drawing, making it currently unsuitable for real-time industrial throughput.

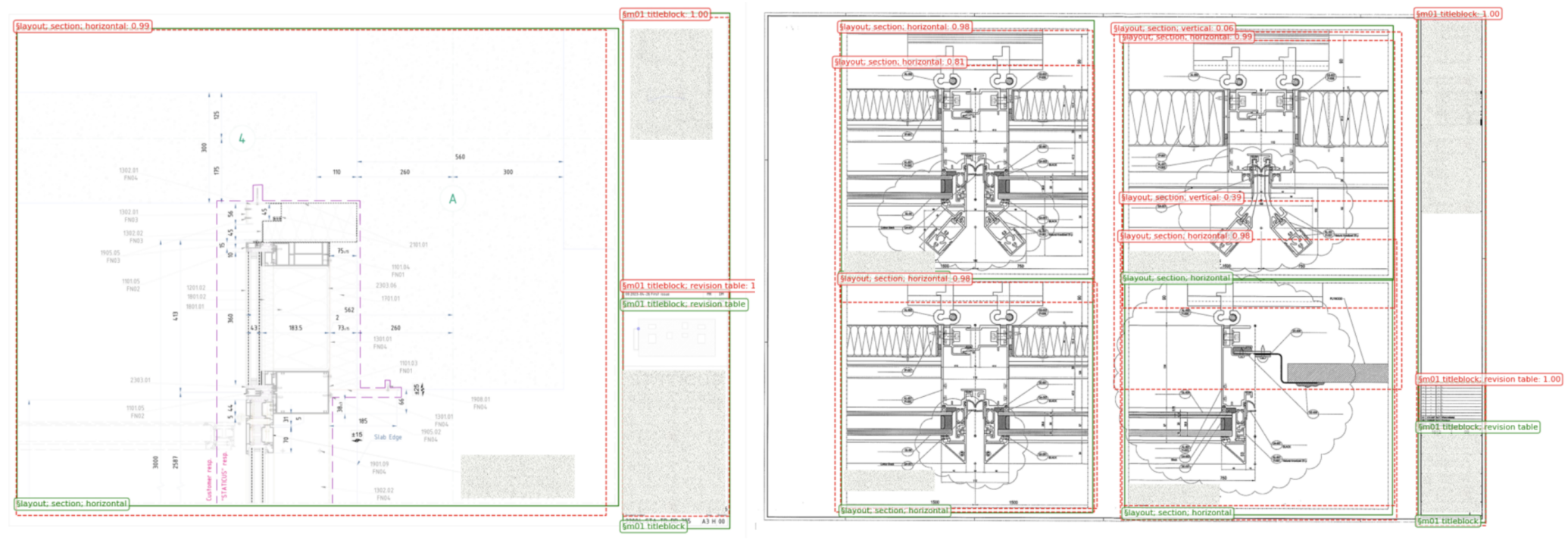

*Figure 3: Visualization of Qwen3-VL inference results. The* ***left*** *panel shows a successful layout detection; the* ***right*** *panel illustrates a failed case with redundant overlaps.*

Finally, DocLayout-YOLO, pre-trained on DocLayNet and DocSynth, fails to outperform general-purpose vision models on the FELD test set. This outcome indicates that architectural innovations and pre-trained features derived from general DLA do not effectively transfer to the AEC domain. The distinctive topological logic and the high-density, line-based graphical-symbolic logic of architectural drawings remain fundamentally different from the structural patterns found in general documents. A more granular analysis of this domain gap will be conducted in the Ablation Study section.

### Category-specific Analysis

Table 3 provides a granular breakdown of the $mAP_{50:95}$ scores across nine layout categories, revealing distinct performance characteristics of the evaluated models.

First, RF-DETR (m) exhibits a dominant performance by achieving the highest precision across all categories. Its consistent lead validates the robustness of the Transformer-based architecture in managing diverse layout structures. Meanwhile, Faster R-CNN and YOLOv12 (m) alternate as the second-best performers depending on the category. Faster R-CNN maintains a competitive edge in regular geometric areas such as *GA section* (0.820), while YOLOv12 performs reliably in *Titleblock* (0.952) and *Legend* (0.910).

Second, DocLayout-YOLO$^{\dagger}$ shows highly polarized results across different classes. While it achieves respectable performance in structured regions like *Titleblock* (0.894) and *Legend* (0.720), its accuracy collapses in *Notes* (0.395) and *GA section* (0.039). This disparity suggests that while its DLA-oriented architecture can recognize common document-like blocks, it struggles with the specialized graphical density and topological complexity inherent in AEC-specific sections.

Despite mAP calculation disadvantages, Qwen3-VL-8B achieved a perfect 1.000 $mAP_{50:95}$ for *Legends* and led in *Vertical Section* (0.856). While highly competitive in *Titleblocks* (0.939), its performance in *Notes* (0.560) was hampered by over-generalization, frequently misidentifying miscellaneous text as notes. This underscores the model's robust pattern recognition alongside current limitations in fine-grained textual classification.

Finally, a cross-model observation reveals that *GA section* and *Plan* generally yield the lowest scores among all categories. This trend is primarily attributed to the limited sample size for these specific classes in the current dataset, which restricts the models' ability to learn sufficient discriminative features during the training phase.

### Ablation Study on DocLayout-YOLO

To further investigate the impact of architectural innovations and pre-training weights of DocLayout-YOLO on model performance within the FELD dataset, we conducted a comprehensive ablation study.

DocLayout-YOLO is a variant of the YOLOv10m baseline. Its primary structural modification involves replacing the standard C2f and C2fCIB modules in layers 4, 6, and 8 with specialized G2L_CRM modules designed for DLA. Additionally, the standard DocLayout-YOLO leverages large-scale pre-training on general document layout datasets, specifically DocLayNet and DocSynth. We designed this ablation study to specifically examine how these DLA-centric modifications and pre-training weights respond to the unique graphical nature of AEC drawings.

We evaluated three specific configurations to isolate these variables:

1. YOLOv10m (COCO pre-trained): The baseline general-purpose object detector with COCO weights.
2. DocLayout-YOLO (COCO + Random Init): Incorporates the G2L_CRM architectural changes but utilizes COCO weights for common layers and random initialization for the new modules, effectively testing the architecture without DLA-specific pre-training.
3. DocLayout-YOLO (DocLayNet&DocSynth pre-trained): The full model incorporating both architectural changes and large-scale document-domain pre-training.

The results presented in Table 4 revealed a surprising trend: both the architectural innovations and the document-specific pre-training weights of DocLayout-YOLO resulted in a performance degradation on the AEC dataset compared to the baseline.

Moving from the YOLOv10m baseline to the DocLayout-YOLO architecture (COCO + Random Init) led to a de-crease in $mAP_{50:95}$ from 0.772 to 0.727. This suggests that while G2L_CRM modules may excel at hierarchical text layouts, they are less effective at capturing the high-density, purely graphical symbolic logic of architectural drawings. While the most significant drop occurred when applying the DocLayNet pre-trained weights, which caused the $mAP_{50:95}$ to plummet to 0.589.

*Table 4: Ablation Study on Model Architecture and Pre-training Weights of DocLayout-YOLO*

| Model / Configuration | $mAP_{50:95}$ | $mAP_{50}$ | P | R | F1 | Params |
|---|---|---|---|---|---|---|
| DocLayout-YOLO (DocLayNet&DocSynth Pre-trained) | 0.589 | 0.675 | 0.736 | 0.602 | 0.662 | 19.97M |
| DocLayout-YOLO (COCO + Random Init) | <u>0.727</u> | <u>0.854</u> | <u>0.921</u> | <u>0.695</u> | <u>0.792</u> | 19.97M |
| YOLOv10m (COCO pre-trained) | **0.772** | **0.900** | **0.954** | **0.826** | **0.885** | 16.46M |

This confirms a strong "domain interference" effect, indicating that the structural priors learned from general documents like newspapers and academic papers are fundamentally incompatible with AEC drawings. The features optimized for general documents act as a negative prior, preventing the model from adapting to the complex graphical logic of architectural drawings.

## Conclusion and future work

This research identifies and addresses a critical "missing link" in the automated interpretation of AEC engineering drawings: layout detection. We systematically evaluated the performance of five deep learning architectures on this task, revealing how different modeling paradigms navigate the unique structural constraints of AEC layouts.

Our experimental results demonstrate that RF-DETR (m) is the most effective technical pathway for this task, achieving a superior $mAP_{50}$. The success of RF-DETR highlights the necessity of global attention mechanisms in capturing the intricate spatial dependencies inherent in large-format engineering drawings. Notably, Qwen3-VL-8B also performed excellently, achieving the highest F1-score, which demonstrates the significant potential of VLMs for interpreting AEC engineering drawing. Conversely, the benchmarking of DocLayout-YOLO revealed a significant "domain gap". Our ablation study specifically proved that architectural priors and pre-trained weights optimized for text-centric documents can act as a negative prior, hindering the model's ability to adapt to the high-density graphical-symbolic logic of AEC layouts.

Despite these advancements, the FELD dataset's limited scale and facade-specific focus restrict broader generalisability. To address this, future work will enhance robustness via domain-specific augmentation and evaluate cross-domain generalisation through a dual-metric framework: quantifying "domain shift" via zero-shot testing and measuring "adaptation velocity" through few-shot transfer learning across structural and MEP sectors.

Furthermore, the inherent high-resolution nature of AEC drawings poses a significant challenge for standard deep learning pipelines; our current methodology simply relies on standard resizing, which inevitably leads to the loss of fine-grained textual features and thin geometric lines. This factor likely constrained the localization precision of RF-DETR and prevented Qwen3-VL from fully leveraging its native capacity for dynamic, native-resolution visual inputs.

Future research efforts will focus on three primary dimensions to achieve more holistic drawing understanding. First, we aim to expand the FELD dataset into a multi-domain AEC layout benchmark to enhance cross-domain generalization and address current class imbalances. Second, in this work, ADIRO is used primarily as a layout-centric schema to support consistent dataset construction and annotation, rather than as a full reasoning layer; in future work, we plan to integrate ADIRO more deeply into the pipeline by leveraging its structure in model design and training, and by using it as a foundation for ontology-based reasoning, knowledge graph construction, and potential integration with BIM models. Finally, we will leverage recognized layout boundaries as "middleware" to guide Main Drawing Content Layer extraction, implementing semantic-aware tiling to ensure that downstream symbol recognition and geometric analysis are performed within a coherent logical context.

## References

Bai, S., Cai, Y., Chen, R., et al. (2025). Qwen3-VL technical report. arXiv preprint arXiv:2511.21631.

Carrara, A., Nousias, S., and Borrmann, A. (2025). Content-based classification of construction drawings. In Proceedings of the 32nd EG-ICE International Workshop on Intelligent Computing in Engineering, Glasgow, UK.

Chen, Y., Liu, R., Zheng, J., et al. (2025). Graph-based document structure analysis. In The Thirteenth International Conference on Learning Representations (ICLR).

Heras, L.-P., Ahmed, S., Liwicki, M., et al. (2014). Statistical segmentation and structural recognition for floor plan interpretation. International Journal on Document Analysis and Recognition, 17(3):221–237.

Huang, Y., Lv, T., Cui, L., et al. (2022). LayoutLMv3: Pre-training for document AI with unified text and image masking. In Proceedings of the 30th ACM International Conference on Multimedia, pages 4083–4091. ACM.

Jamieson, L. (2024). Deep Learning for Digitising Complex Engineering Drawings. Phd thesis, Robert Gordon University.

Jamieson, L., Moreno-Garcia, C. F., and Elyan, E. (2025). Towards fully automated processing and analysis of construction diagrams: AI-powered symbol detection. International Journal on Document Analysis and Recognition, 28(1):71–84.

Kashevnik, A., Shilov, N., Teslya, N., et al. (2023). An approach to engineering drawing organization: Title block detection and processing. IEEE Access, 11:15143–15155.

Khan, M. T., Chen, L., Ng, Y. H., et al. (2024). Fine-tuning vision-language model for automated engineering drawing information extraction. arXiv preprint arXiv:2411.03707.

Khan, M. T., Yong, Z., Chen, L., et al. (2026). A multi-stage hybrid framework for automated interpretation of multi-view engineering drawings using vision language model. In Proceedings of the 13th International Conference on Industrial Engineering and Applications (ICIEA).

Kim, G., Hong, T., Yim, M., et al. (2022). OCR-free document understanding transformer. In Computer Vision – ECCV 2022, volume 13688 of Lecture Notes in Computer Science, pages 498–517. Springer.

Lombardi, A., Duan, L., Elnagar, A., et al. (2025). Title block detection and information extraction for enhanced building drawings search. In 2025 European Conference on Computing in Construction (EC3).

Namboodiri, A. M. and Jain, A. K. (2007). Document Structure and Layout Analysis, pages 29–48. Springer London.

Pfitzmann, B., Auer, C., Dolfi, M., et al. (2022). DocLayNet: A large human-annotated dataset for document-layout segmentation. In Proceedings of the 28th ACM SIGKDD Conference on Knowledge Discovery and Data Mining, pages 3743–3751. ACM.

Ren, S., He, K., Girshick, R., et al. (2017). Faster R-CNN: Towards real-time object detection with region proposal networks. IEEE Transactions on Pattern Analysis and Machine Intelligence, 39(6):1137–1149.

Robinson, I., Robicheaux, P., Popov, M., et al. (2026). RF-DETR: Neural architecture search for real-time detection transformers. In The Fourteenth International Conference on Learning Representations (ICLR).

Schönfelder, P. and König, M. (2025). Ontology-based reasoning in automatic floor plan analysis. Advanced Engineering Informatics, 68:103761.

Sechidis, K., Tsoumakas, G., and Vlahavas, I. (2011). On the stratification of multi-label data. In Machine Learning and Knowledge Discovery in Databases (ECML PKDD 2011), volume 6913 of Lecture Notes in Computer Science, pages 145–158. Springer.

Tian, Y., Ye, Q., and Doermann, D. (2025). Yolov12: Attention-centric real-time object detectors. In Advances in Neural Information Processing Systems 38 (NeurIPS).

Wang, Z., Xu, Y., Cui, L., et al. (2021). LayoutReader: Pre-training of text and layout for reading order detection. In Proceedings of the 2021 Conference on Empirical Methods in Natural Language Processing, pages 4735–4744. Association for Computational Linguistics.

Xu, Y., Li, M., Cui, L., et al. (2020). LayoutLM: Pre-training of text and layout for document image understanding. In Proceedings of the 26th ACM SIGKDD International Conference on Knowledge Discovery and Data Mining, pages 1192–1200. ACM.

Xu, Y., Xu, Y., Lv, T., et al. (2021). LayoutLMv2: Multi-modal pre-training for visually-rich document understanding. In Proceedings of the 59th Annual Meeting of the Association for Computational Linguistics and the 11th International Joint Conference on Natural Language Processing (Volume 1: Long Papers), pages 2579–2591. Association for Computational Linguistics.

Zhang, W., Joseph, J., Yin, Y., et al. (2023). Component segmentation of engineering drawings using graph convolutional networks. Computers in Industry, 147:103885.

Zhao, Y., Huang, J., Hu, J., et al. (2024a). SWIFT: A scalable lightweight infrastructure for fine-tuning. arXiv preprint arXiv:2408.05517.

Zhao, Z., Kang, H., Wang, B., et al. (2024b). DocLayout-YOLO: Enhancing document layout analysis through diverse synthetic data and global-to-local adaptive perception. arXiv preprint arXiv:2410.12628.

Zhong, X., Tang, J., and Yepes, A. J. (2019). PubLayNet: Largest dataset ever for document layout analysis. In 2019 International Conference on Document Analysis and Recognition (ICDAR), pages 1015–1022.